\documentclass[accepted,mathfont=cm]{uai2021} % Computer Modern math instead of
\usepackage{amsmath,amsfonts,bm}

\def\1{\bm{1}}

\def\vs{{\bm{s}}}

\def\mA{{\bm{A}}}
\def\mB{{\bm{B}}}

\def\mD{{\bm{D}}}

\def\mF{{\bm{F}}}

\def\mO{{\bm{O}}}
\def\mP{{\bm{P}}}
\def\mQ{{\bm{Q}}}
\def\mR{{\bm{R}}}
\def\mS{{\bm{S}}}

\def\mU{{\bm{U}}}

\def\mW{{\bm{W}}}
\def\mX{{\bm{X}}}

\DeclareMathAlphabet{\mathsfit}{\encodingdefault}{\sfdefault}{m}{sl}
\SetMathAlphabet{\mathsfit}{bold}{\encodingdefault}{\sfdefault}{bx}{n}

\def\emA{{A}}
\def\emB{{B}}

\def\emF{{F}}

\def\emO{{O}}

\def\emS{{S}}

\def\emW{{W}}

\newcommand{\R}{\mathbb{R}}

\usepackage[american]{babel}
\usepackage{natbib} % has a nice set of citation styles and commands
\usepackage{mathtools} % amsmath with fixes and additions
\usepackage{booktabs} % commands to create good-looking tables
\usepackage{tikz} % nice language for creating drawings and diagrams
\usepackage{overpic}
\usepackage{amssymb}
\usepackage{graphicx}
\usepackage{subfigure}
\usepackage{arydshln}
\usepackage{enumitem}
\usepackage{makecell}
\usepackage[ruled,norelsize]{algorithm2e}
\usepackage{listings}
\usepackage{color, xcolor, colortbl}

\usepackage{hyperref}
\usepackage{cleveref}
\crefname{equation}{Eq.}{Eq.}
\crefname{figure}{Fig.}{Fig.}
\crefname{table}{Tab.}{Table~}
\crefname{section}{Sec.}{Section~}
\crefname{algorithm}{Alg.}{Algorithm~}

\title{Structured Sparsification with Joint Optimization of \\
Group Convolution and Channel Shuffle}

\newcommand*\samethanks[1][\value{footnote}]{\footnotemark[#1]}

\author[1]{{Xin-Yu Zhang\thanks{Xin-Yu Zhang (\href{mailto:Xinyu Zhang <xinyuzhang@mail.nankai.edu.cn>?Subject=Structured sparsification paper}{xinyuzhang@mail.nankai.edu.cn}) and Kai Zhao contribute equally to this work.}}} % Lead author
\author[1]{Kai Zhao\samethanks}
\author[2]{Taihong Xiao}
\author[1]{Ming-Ming Cheng}
\author[2]{Ming-Hsuan Yang}
\affil[1]{%
    TKLNDST\\
    CS\\
    Nankai University
}
\affil[2]{%
    University of California, Merced
}

\graphicspath{{./imgs/}}
\DeclareGraphicsExtensions{.pdf,.jpg,.png}

\def\name{StrucSpars}
\def\vs{\textit{vs.}}
\def\ie{\textit{i.e.,~}}
\def\eg{\textit{e.g.,~}}

\def\vs{{\em v.s.}}

\begin{document}
\maketitle

\begin{abstract}
    Recent advances in convolutional neural networks (CNNs)
    usually come with the expense of excessive computational overhead and memory footprint.
    Network compression aims to alleviate this issue by training compact
    models with comparable performance.
    However, existing compression techniques either entail dedicated expert design
    or compromise with a moderate performance drop.
    In this paper, we propose a novel structured sparsification
    method for efficient network compression.
    The proposed method automatically induces structured sparsity % structurally sparse representations
    on the convolutional weights, thereby facilitating
    the implementation of the compressed model with the highly-optimized group convolution.
    % the compressed model to be easily implemented with the
    % highly-optimized group convolution.
    %
    We further address the problem of inter-group communication
    with a learnable channel shuffle mechanism.
    The proposed approach can be easily applied to compress many network architectures with a negligible performance drop.
    Extensive experimental results and analysis demonstrate that our approach
    gives a competitive performance against the recent network compression counterparts with a sound accuracy-complexity trade-off.
\end{abstract}

\section{Introduction}

Convolutional Neural Networks (CNNs) have made significant advances in a wide
range of vision and learning tasks \citep{ILSVRC15,long2015fully}.
However, the performance gains usually entail a heavy computational cost,
which makes the deployment of CNNs on portable devices difficult. 
To meet the memory and computational constraints in real-world applications,
numerous model compression techniques have been developed. 

Existing network compression techniques are mostly based on weight
quantization \citep{chen2015compressing,courbariaux2016binarized,rastegari2016xnor,wu2016quantized},
knowledge distillation \citep{hinton2015distilling,chen2017learning,yim2017gift},
or network pruning \citep{li2017pruning,he2017channel,liu2017learning,molchanov2019importance,tang2020scop}.
Weight quantization methods use low bit-width numbers to represent weights and activations,
which usually bring a moderate performance degradation.
Knowledge distillation schemes transfer knowledge from a large
teacher network to a compact student network, which are typically susceptible
to the teacher/student network architectures \citep{mirzadeh2019improved,liu2019search}.
Closely related to our work, network pruning approaches reduce the model
size by removing a proportion of model parameters that are considered unimportant.
Notably, filter pruning algorithms \citep{li2017pruning,liu2017learning,molchanov2019importance,tang2020scop}
remove the entire filters and result in structured architectures that can be
readily incorporated into modern BLAS libraries.

Identifying unimportant filters is critical to the filter pruning methods. 
It is well-known that the weight norm can serve as a good indicator
of the corresponding filter importance \citep{li2017pruning,liu2017learning}.
Filters corresponding to smaller weight norms are considered to
contribute less to the outputs.
Furthermore, 
the $L_1$ regularization can be used to promote sparsity
\citep{liu2017learning}.
However, there are still several issues in the existing pruning methods:
1) pruning a large proportion of convolutional filters will result in
severe performance degradation;
2) pruning alters the input/output feature dimensions, and thus meticulous 
adaptation is required to handle network architectures with shortcut connections
(\eg residual connections \citep{he2016deep} and dense connections
\citep{huang2017densely}).

Before presenting the proposed method, we briefly introduce the
group convolution (GroupConv) \citep{ILSVRC15},
which plays an important role in this work.
For the vallina convolution operation, the output features are
densely-connected with the input features,
while for the GroupConv, the input features are equally split
into several groups and transformed within each group independently.
Essentially, each output channel is connected with only a
proportion of the input channels, which leads to sparse neuron connections.
Therefore, deep CNNs with GroupConvs can be trained on less
powerful GPUs with smaller amount of memory. 
%
%Back to the network compression literature, 
%The basic assumption of network compression methods is that
% removing unimportant neuron connections and inducing sparse neuron connectivity
% can better preserve the representational capacity of the compressed model
% compared with pruning the entire filters.
%
% Viewed in the network compression literature, removing a proportion of
% unimportant neuron connections instead of pruning the entire filters may
% assist preserve the representational capacity of the compressed model.

% In this work, we propose a fourth solution towards efficient network
% compression, namely, \textit{structured sparsification},
In this work, we propose a novel approach for network compression
where unimportant neuron connections are pruned to facilitate the usage of GroupConvs.
Nevertheless, converting vallina convolutions into GroupConvs is a challenging task.
%
%MH: valid GroupConvs? What do you mean? never mind. I understand it now.
First, not all sparse neuron connectivities correspond to valid GroupConvs,
while certain requirements must be satisfied, \eg mutual exclusiveness
of different groups.
To guarantee the desired structured sparsity,
we impose \textit{structured regularization} upon the convolutional weights
and zero out the sparsified weights.
Another challenge is that stacking multiple GroupConvs sequentially will hinder
the inter-group information flow.
\citet{zhang2018shufflenet} propose the ShuffleNet with a \textit{channel shuffle} mechanism,
\ie gathering channels from distinct groups, to ensure the inter-group communication,
though the order of permutation is hand-crafted.
However, we solve the problem of channel shuffle in a learning-based scheme.
% Different from \cite{zhang2018shufflenet}, however,
% the optimal channel permutation is automatically searched
% over every possible permutation in our approach.
%
Concretely, we formulate the learning of channel shuffle
as a linear programming problem, which can be solved by efficient
algorithms like the network simplex method \citep{bonneel2011displacement}.
% Due to the difficulty of minimization over the set of permutation matrices,
% we relax the feasible space and solve the problem via linear programming.
%
Since the structured sparsity is induced among the convolutional
weights, our method is nominated as
\textit{structured sparsification}.
% , abbreviated to \textbf{SS}.

%MH: check efficient
% In this work, we propose the \textit{structured sparsification} method
% for efficient and effective network compression.
%
The proposed structured sparsification method is designed for three goals. 
(a) Our approach can better handle those network architectures with shortcut connections.
A wide range of backbone architectures are amenable to
our method without the need for any special adaptation.
(b) Our method is capable of achieving \textit{high compression rates}.
In modern efficient network architectures, the complexity of $3 \times 3$ 
convolutions is highly compressed, while the computation bottleneck becomes
the point-wise convolutions (\ie $1 \times 1$ convolutions) \citep{zhang2018shufflenet}.
For example, the point-wise convolutions occupy 81.5\% of the total
FLOPs in the MobileNet-V2 \citep{sandler2018mobilenetv2} backbone and
93.4\% in ResNeXt \citep{xie2017aggregated}.
Our method is applicable to all convolution operators
so that a high compression rate is reachable.
% the empirical compression rate can be significantly higher
% than the concurrent network compression methods.
%
(c) Our approach brings \textit{negligible performance drop}.
As all filters are preserved under our methodology,
we retain stronger representational capacity of the compressed model
and achieve better accuracy-complexity trade-off
than the pruning-based counterparts (see \cref{fig:trade-off}).
% The merits of the structured sparsification are three-fold:
% \begin{enumerate}[label=(\roman*)]
%     \item \textbf{Model-agnostic}: a wide range of backbone architectures
%     (\eg ResNet \cite{he2016deep} and DenseNet \cite{huang2017densely})
%     are amenable to our method without the need of any special adaptation;

%     \item \textbf{Highly-compressible}: our approach is applicable to all
%     convolution operators so that the empirical compression rate can be
%     significantly higher that the concurrent network compression methods,
%     \eg on the ResNet-50 backbone, an accuracy drop of 4.63\% on the
%     ImageNet \cite{ILSVRC15} benchmark is sacrificed for an $\sim$80\% compression;

%     \item \textbf{Bringing Negligible performance drop}:
%     our method achieves state-of-the-art accuracy-efficiency trade-off in favor of
%     all pruning-based counterparts and brings minimal performance degradation,
%     \eg on the ResNet-101 backbone, a negligible accuracy drop of 0.48\% is
%     is sacrificed for a $\sim$40\% compression.
% \end{enumerate}

The main contributions of this work are as follows.
\begin{itemize}
    \item We propose a \textit{learnable channel shuffle mechanism} in which the permutation of the convolutional weight norm matrix is learned via linear programming;
    
    \item We formulate a novel \textit{structured sparsification} framework for efficient network compression, which unifies weight pruning, GroupConv, and the learnable channel shuffle;
    
    \item The experimental results on the CIFAR-10 and ImageNet datasets demonstrate that the proposed structured sparsification performs well against the concurrent filter pruning approaches with a balanced trade-off of  accuracy and complexity.
\end{itemize}

%\begin{compactitem}
%    \item 
%    \item 
    %  Upon the permuted weight norm matrix, we impose
    % \textit{structured regularization} (\cref{sec:sparsification})
    % to obtain valid GroupConvs by zeroing out the sparsified weights.
%\item 
%MH: How many times do I need to tell you not to state this way?
%Experimental results on the CIFAR and ImageNet datasets demonstrate thatthe proposed structured sparsification achieves a better accuracy-complexity   trade-off than state-of-the-art filter pruning approaches.
%
    % With the structurally sparse convolutional weights,
    % we design the \textit{criteria of learning cardinality}
    % (\cref{sec:criteria}) in which unimportant neuron connections are pruned
    % with minimal impact on the entire network.
%\end{compactitem}
%
% Incorporating the learnable channel shuffle mechanism,
% the structured regularization and the grouping criteria,
% the proposed structured sparsification method
% performs favorably against the state-of-the-art
% network pruning techniques on both CIFAR \cite{krizhevsky2009learning}
% and ImageNet \cite{ILSVRC15} datasets.

\section{Related Work}

\paragraph{Network Compression.}
Compression methods for deep models can be broadly categorized based on \textit{weight quantization},
\textit{knowledge distillation}, or \textit{network pruning}.
Closely related to our work are the network pruning approaches based on \textit{filter pruning}.
It is well-acknowledged that filters with smaller weight norms are considered
to make negligible contribution to the outputs and can be pruned.
\citet{li2017pruning} prune filters according to the $L_1$ norm of the convolutional weights,
while \citet{liu2017learning} prune models with batch normalization
\citep{ioffe2015batchbatch} by removing the channels with smaller batch-norm
scaling factors.
An $L_1$ regularization term is further imposed on these scaling factors
to promote sparsity.
% \xth{I cannot see the difference between these two methods. Could you please rephrase the sentence. Doe the first one impose any constraint for sparsity before pruning?}
%Xinyu: The main difference is that the former consider the norm of convolutional weights as the indicator of importance while the latter consider the norm of the BN scaling factor (conv norm vs BN factor).

% Different from filter pruning, our proposed
% structured sparsification promotes the structured sparsity
% among neuron connections between consecutive hidden layers instead of removing the entire filters.
% %
% We find that removing the entire filters may impair the
% representational capacity significantly and result in unbalanced
% \xth{(why unbalanced? Needs more explanation.)} pruned architectures.
% %
% On the contrary, our approach merely removes certain unimportant
% connections and reserves the entire filters. 
% %
% In this way, the capacity of the network is less reduced compared
% with the pruning-based approaches.
% %
% Moreover, our method do not alter the input/output dimensions,
% thus can be incorporated into a wide range of backbone architectures with ease.
% 
However, removing those filters corresponding to smaller weight
norms may significantly reduce the representational capacity.
Instead, we propose a structured sparsification method that enforces structured sparsity among neuron connections
and merely removes certain unimportant connections
while the entire filters are preserved.
As a result, the network capacity is less affected than the pruning-based approaches \citep{li2017pruning,liu2017learning,molchanov2019importance,tang2020scop}.
Furthermore, our method does not alter the input/output dimensions,
and can be easily incorporated into numerous backbones.

\paragraph{Group Convolution.}
Group convolution (GroupConv) is introduced in the AlexNet \citep{ILSVRC15}
to overcome the GPU memory constraints. 
GroupConv partitions the input features into
mutually exclusive groups and transforms the features within each
group in parallel.
Compared with the vallina (\ie densely connected) convolution,
a GroupConv with $G$ groups can reduce the computational cost and
number of parameters by a factor of $G$.
The ResNeXt \citep{xie2017aggregated} designs a multi-branch architecture
by employing GroupConvs and defines the \textit{cardinality} as the number of
parallel transformations, which is simply the group number in each GroupConv.
If the cardinality equals to the number of channels,
GroupConv becomes the \textit{depthwise separable convolution},
which is widely used in recent lightweight neural architectures
\citep{howard2017mobilenets,sandler2018mobilenetv2,zhang2018shufflenet,
ma2018shufflenet,chollet2017xception}.

However, the aforementioned methods all treat the cardinality as a hyper-parameter,
and the connectivity patterns between consecutive features are hand-crafted as well.
On the other hand, there is also a line of research focusing on learnable GroupConvs
\citep{huang2018condensenet,wang2019fully,zhang2019differentiable}.
Both CondenseNet \citep{huang2018condensenet} and FLGC \citep{wang2019fully} 
pre-define the cardinalities of GroupConvs and learn the connectivity patterns.
%
% Notably, \cite{huang2018condensenet} introduces an \textit{index layer},
% in which features are selected and re-arranged so that regular GroupConvs can
% be readily applied.
%
We note that the work by \citet{zhang2019differentiable} learns the cardinality and neuron connectivity simultaneously.
Essentially, this dynamic grouping convolution
is modeled by a binary relationship matrix $\mU$ where
$U_{ji}$ indicates the connectivity between the $i^{th}$ input
channel and the $j^{th}$ output channel.
To guarantee that the resulting operator is a valid GroupConv,
the relationship matrix is constructed using a Kronecker product
of several binary symmetric $2 \times 2$ matrices, which is a sufficient but unnecessary condition. Consequently, the space of all valid GroupConvs is not fully exploited.

In contrast to the prior art \citep{zhang2019differentiable}, our method can learn the optimal connection over all possible neuron connectivity patterns, thus resulting in better model structures and performance.
The connection between two layers is represented by the composition of a permutation matrix (over the permutation set) and the convolutional weights.
The importance of each neuron connection is quantified by the corresponding entry of the weight norm matrix.

% We quantify the importance of each neuron connection by the structure sparsity regularization on the permuted weight norm matrix.

% Our method decouples the learning of cardinality and connectivity.
% %
% Motivated by the norm-based criterion in the network pruning methods
% \cite{li2017pruning,liu2017learning},
% we quantify the importance of each neuron connection by its weight norm and learn the connectivity pattern over the permutation set imposed on the weight norm matrix.
% %
% Besides, the structured regularization is imposed on the permuted weight norm matrix and the cardinality is learned accordingly.
% %
% The essential difference between our approach and prior art
% \cite{zhang2019differentiable} is that
% all possible neuron connectivity patterns, \ie relationship
% matrices, can be reached by our method.

\paragraph{Channel Shuffle Mechanism.}
The ShuffleNet \citep{zhang2018shufflenet} combines the channel shuffle
mechanism with GroupConv for efficient network design, in which channels
from different groups are gathered so as to facilitate the inter-group communication.
Without channel shuffle, stacking multiple GroupConvs will eliminate the information flow among different groups and weaken the representational
capacity.
Different from the hand-crafted counterpart 
\citep{zhang2018shufflenet}, the proposed channel shuffle operation is learnable over the space of all possible channel permutations.
Furthermore, without bells and whistles, our channel shuffle only involves a simple permutation along the channel dimension, which can be conveniently implemented by an index operation.

\paragraph{Neural Architecture Search.}
%While the state-of-the-art CNNs
%\cite{vgg,he2016deep,xie2017aggregated,huang2017densely} are mostly manually designed, 
Neural Architecture Search (NAS) \citep{zoph2016neural,baker2016designing,zoph2018learning,real2019regularized,wu2019fbnet}
aims to automate the process of designing neural architectures within certain budgets of computational resources.
% Numerous Neural Architecture Search (NAS) methods 
% \cite{zoph2016neural,baker2016designing,zoph2018learning,
% real2019regularized,wu2019fbnet} based on
%MH: do no use popular... it is not important whether a method is porpular or not..
%Popular algorithms include reinforcement learning
Existing NAS algorithms are developed based on reinforcement learning
\citep{zoph2016neural,baker2016designing,zoph2018learning},
evolutionary search \citep{real2019regularized},
and differentiable approaches \citep{liu2018darts,wu2019fbnet}.
% have been developed in recent years. 
%
Our method can be viewed as a special case of hyper-parameter
(\ie cardinality) optimization and neuron connectivity search.
However, different from existing approaches evaluated on numerous architectures,
the proposed method can determine the compressed architecture in one single training pass.
We highlight that the efficiency of our approach is in accordance with the aim
of neural architecture search.
% \xth{Pay attention to your claim. Is determining the compressed architecture in one single training pass related to the scalability?}
% to significantly reduce the computational load in terms of time and memory. 
%
%As such, the training time and computational
%sources put into neural architecture search.

\section{Structured Sparsification}

\subsection{Overview} \label{sec:overview}

The structured sparsification method is designed to
zero out a proportion of the convolutional weights so that the vanilla convolutions can be structured into group convolutions (GroupConvs) via a learned permutation matrix.
We adopt the ``train, compress, finetune'' pipeline, 
in a way similar to the recent pruning approaches \citep{liu2017learning}.
Concretely, we first train a large model under the structured regularization,
then compress vanilla convolutions into GroupConvs under a certain criteria,
and finally finetune the compressed model to recover accuracy.
%
% The connectivity patterns can be learned as the structured  regularization heavily depends on them.
%
To this end, three issues need to be addressed: 
1) how to learn the connectivity patterns (\cref{sec:lp});
2) how to design the structured regularization (\cref{sec:sparsification});
3) how to decide the grouping criteria (\cref{sec:criteria}).
%
%The solutions can be found in \cref{sec:lp}, \ref{sec:sparsification},
%and \ref{sec:criteria}, respectively.
%
Additional details of our pipeline are presented in \cref{sec:details}.
% the supplementary materials.

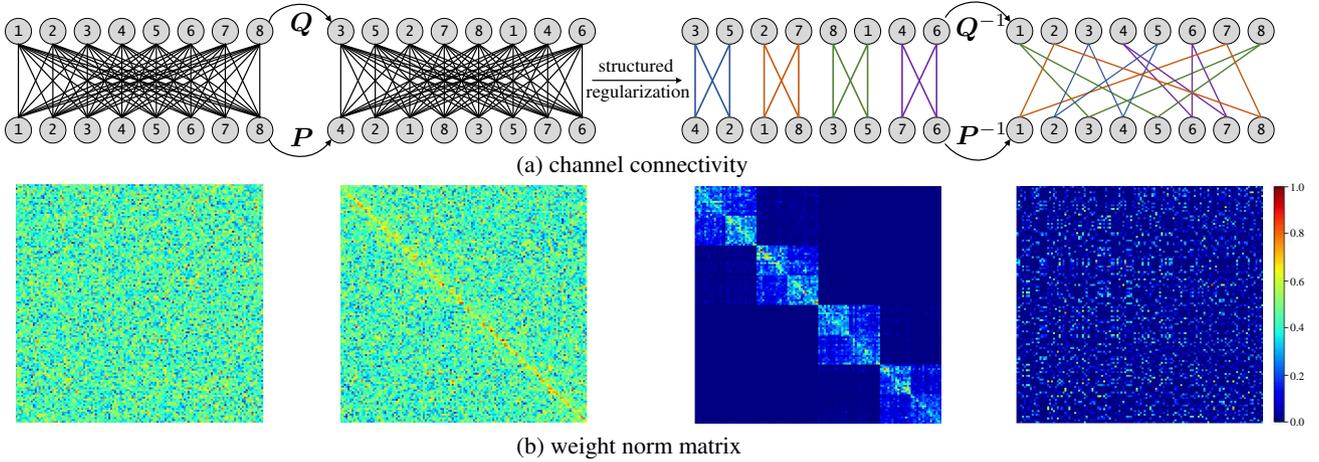
\begin{figure*}[!t]
    \centering
    \begin{overpic}[width=\textwidth]{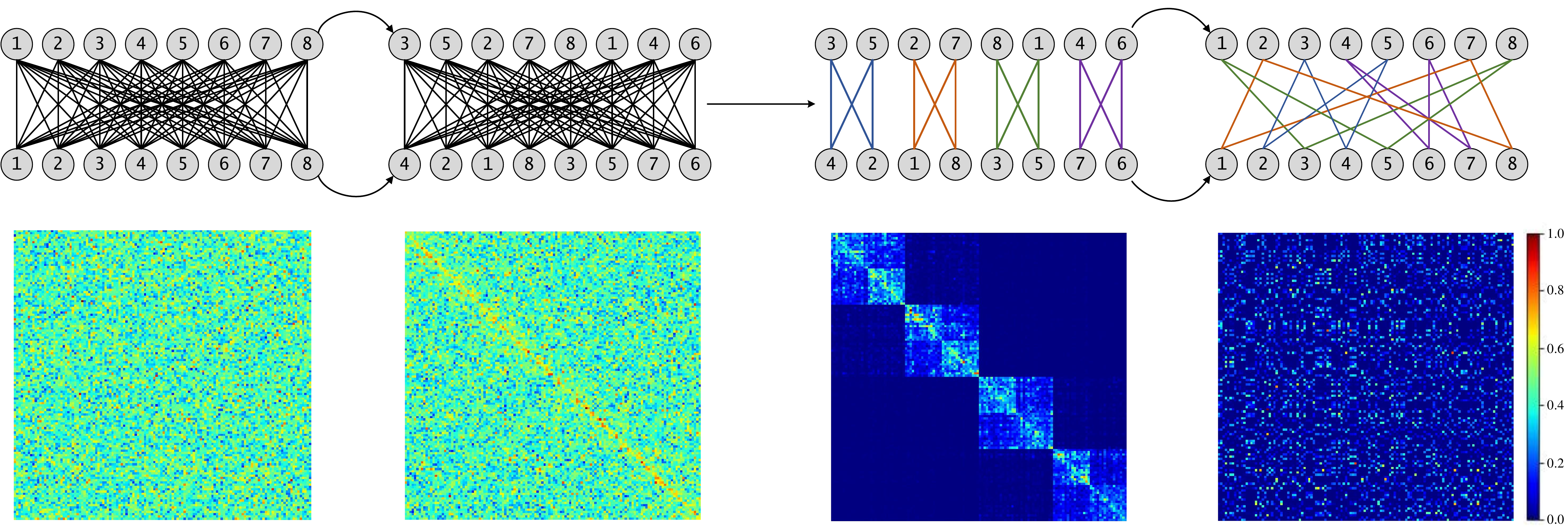}
    \put(21.9,31.1){$\mQ$}
    \put(21.9,22.1){$\mP$}
    \put(73.0,30.7){$\mQ^{-1}$}
    \put(73.0,22.3){$\mP^{-1}$}
    \put(45.6,27.9){\scriptsize{structured}}
    \put(44.65,25.9){\scriptsize{regularization}}
    \put(39.3,20.1){\footnotesize{(a) channel connectivity}}
    \put(39.3,-1.7){\footnotesize{(b) weight norm matrix}}
    \end{overpic}
    \vspace{0.5pt}
    \caption{Illustration of the learnable channel shuffle mechanism.
    The original convolutional weights (first column) are shuffled along the
    input/output channel dimensions in order to solve \cref{eqn:optimization}.
    The structured regularization is imposed upon the permuted weight
    norm matrix (second column) to promote the structured sparsity, and connections
    with small weight norms are discarded (third column).
    In the original ordering of channels, a structurally sparse connectivity pattern is
    learned (fourth column), and notably every valid connectivity pattern
    can be possibly reached in this manner.
    }
    \label{fig:channel-shuffle}
\end{figure*}

\subsection{Learning Connectivity with Linear Programming}
\label{sec:lp}
Let $\mF \in \R^{C^{\text{in}} \times H \times W}$ be the input feature map,
where $C^{\text{in}}$ denotes the number of input channels.
We apply a vallina convolution\footnote{For simplicity, we omit the
bias term from \cref{eqn:convolution}, and assume the
convolution operator is of stride 1
with proper paddings.} with weights
$\mW \in \R^{C^{\text{out}} \times C^{\text{in}} \times K \times K}$ to $\mF$,
\ie $\mO = \mW \ast \mF$, where $\mO \in \R^{C^{\text{out}} \times H \times W}$
with $C^{\text{out}}$ denoting the number of output channels.
Each entry of $\mO$ is a weighted sum of a local patch
of $\mF$, namely,
\begin{equation}
    \emO_{j,p,q} = \sum_{i,k,l} \emW_{j,i,k,l} \emF_{i,p+k,q+l}.
    \label{eqn:convolution}
\end{equation}
In \cref{eqn:convolution}, the $i^{th}$ channel of $\mF$
relates to the $j^{th}$ channel of $\mO$ via weights $\mW_{j,i,:,:}$.
Motivated by the norm-based importance estimation in filter pruning
 \citep{li2017pruning,liu2017learning},
we quantify the importance of the connection between $\mF_i$ and
of $\mO_j$ by $\| \mW_{j,i,:,:} \|$.
Thus, the importance matrix $\mS \in \R^{C^{\text{out}} \times C^{\text{in}}}$
can be defined as the norm along the ``kernel size'' dimensions
of $\mW$, \ie $\emS_{j,i} = \| \mW_{j,i,:,:} \|$.

Next, we extend our formulation to GroupConvs with
cardinality $G$.
A GroupConv can be considered as a convolution with sparse neuron
connectivity, in which only a proportion of input channels is visible to
each output channel.
Without loss of generality, we assume both $C^{\text{in}}$ and $C^{\text{out}}$
are divisible by $G$, and \cref{eqn:convolution} can be adapted as
\begin{equation}
    \emO_{j,p,q} = \sum_{i=(n-1)m+1}^{nm}
    \sum_{k,l} \emW_{j,i,k,l} \emF_{i,p+k,q+l},
    \label{eqn:groupconv}
\end{equation}
where $n = \textrm{ceil}(jG/C^{\text{out}})$ indicates the $j^{th}$ output
channel belongs to the $n^{th}$ group, and $m = C^{\text{in}} / G$ is
the number of input channels within each group.
Clearly, the valid entries of $\mW$ form a block diagonal
matrix with $G$ equally-split blocks at the input/output channel dimensions.
Thus, the GroupConv module requires $C^{\text{in}} C^{\text{out}} K^2/G$ parameters
and $C^{\text{in}} C^{\text{out}} K^{2} HW/G$ FLOPs for processing the feature $\mF$,
and the computational complexity is reduced by a factor of $G$
compared with the vanilla counterpart.

We note that if a vanilla convolution operator can be converted into GroupConv
without affecting its functional property (we call such convolution operators
\textit{groupable}), the convolutional weights $\mW$ must be block diagonal
after certain permutations along the input/output channel dimensions.
Due to the positive definiteness property of the norm, a necessary and sufficient condition of a convolution operator $\mW$ being groupable is that
\begin{equation}
  \begin{split}
    &\exists \, \mP \in \mathcal{P}^{C^{\text{out}}}~\mbox{and}~\mQ \in \mathcal{P}^{C^{\text{in}}}, \\
    &\mbox{s.t.}~\mP \mS \mQ~\mbox{is block diagonal with equally-split blocks},
  \end{split}
  \label{eqn:s-n-condition}
\end{equation}
where $\mathcal{P}^N$ denotes the set of $N \times N$ permutation matrices.
Here, the permutation matrices $\mP$ and $\mQ$ shuffle the channels of
the input and output features, and thus determine the connectivity pattern
between $\mF$ and $\mO$ (see \cref{fig:channel-shuffle}). 

\begin{figure}[!t]
    \centering
    \begin{overpic}[width=0.49\textwidth]{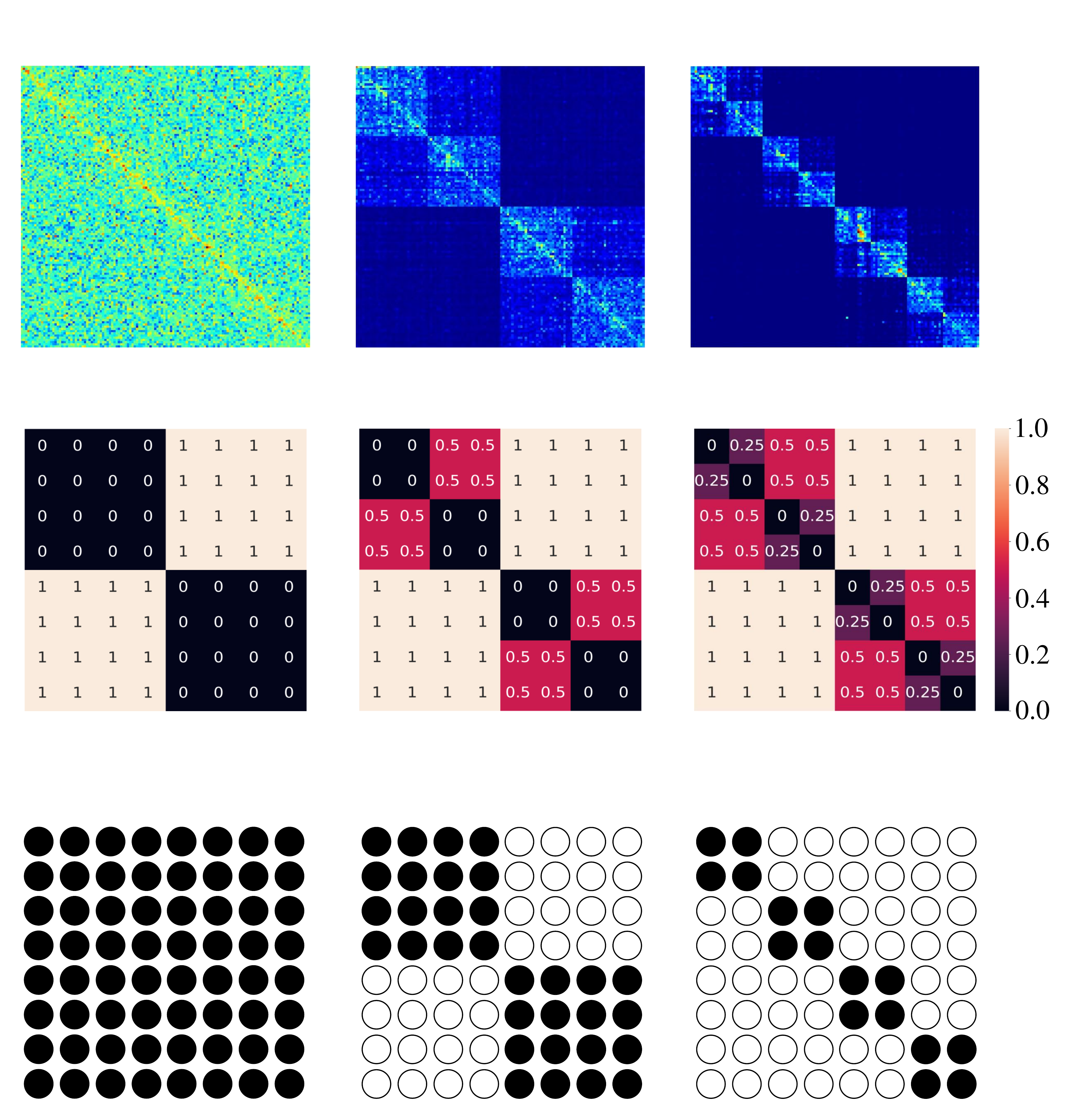}
        \put(3.6,95.4){\footnotesize{group level $=1$}}
        \put(33.5,95.4){\footnotesize{group level $=2$}}
        \put(63.3,95.4){\footnotesize{group level $=3$}}
        \put(16.6,64.4){(a) permuted weight norm matrix $\mS'$}
        \put(13.7,32.7){\footnotesize{$\mR_1$}}
        \put(43.3,32.7){\footnotesize{$\mR_2$}}
        \put(68.6,32.7){\footnotesize{$\mR_3=\mR$}}
        \put(22.2,28.4){(b) structured regularization}
        \put(13.5,-1.7){\footnotesize{$\mU_1$}}
        \put(43.7,-1.7){\footnotesize{$\mU_2$}}
        \put(73.8,-1.7){\footnotesize{$\mU_3$}}
        \put(25.5,-6.4){(c) relationship matrix}
    \end{overpic}
    \vspace{2mm}
    \caption{Illustration of the structured regularization matrix $\mR_g$
    and the relationship matrix $\mU_g$ corresponding to the group level $g$.
    (a) Heat map of the permuted weight norm matrix $\mS'$.
    Non-diagonal blocks of the weight norm are sparsified.
    (b) Structured regularization matrix $\mR_g$.
    The regularization coefficient decays exponentially as the group level grows.
    A special case of the decay rate of 0.5 is demonstrated.
    Besides, the matrix $\mR_g$ depends on the current group level $g$,
    and when the maximal possible group level is achieved, the matrix $\mR_g$
    becomes the cost matrix $\mR$ in \cref{eqn:optimization};
    (c) Relationship matrix $\mU_g$.
    The entries of the permuted weight norm matrix corresponding to the
    zero entries of the relationship matrix will be zeroed out during
    grouping.
    }\label{fig:group-level}
\end{figure}
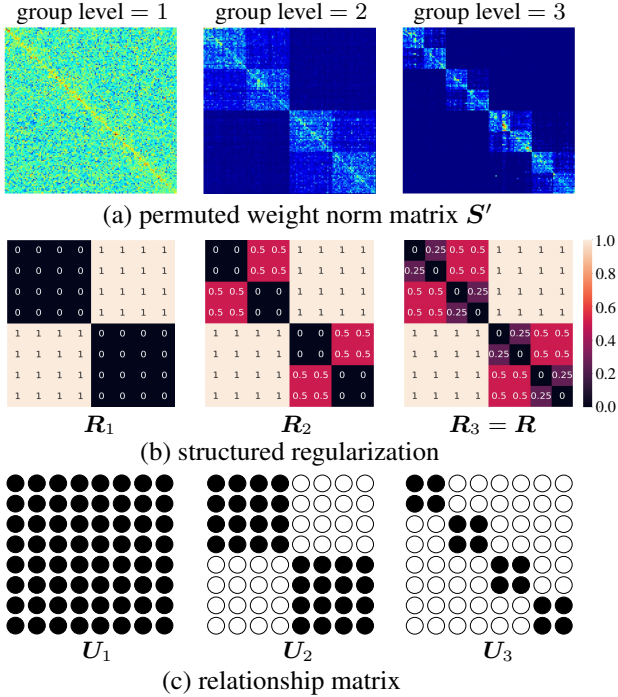

However, a randomly initialized and trained convolution operator
by no means can be groupable unless sparsity constraints are imposed.
To this end, we resort to permuting $\mS$ so as to make $\mS' = \mP \mS \mQ$
``as block diagonal as possible''.
The next question is how to rigorously characterize the term
``as block diagonal as possible''.
Here, we assume both $C^{\text{in}}$ and $C^{\text{out}}$ are powers of
2, where the most widely-used backbone architectures (\eg VGG \citep{vgg} and
ResNet \citep{he2016deep}) satisfy this assumption\footnote{
  Similar reasoning can be applied if both $C^{\text{in}}$ and
  $C^{\text{out}}$ have many factors of 2. See \cref{appendix:reg}.
}.
Then, the potential cardinality is also a power of 2.
As the cardinality grows, more and more non-diagonal blocks are zeroed out
(see \cref{fig:group-level}(c)).
As illustrated in \cref{fig:group-level}(b), we define the cost matrix
$\mR$ to progressively penalize the non-zero entries of the non-diagonal blocks.
Finally, we utilize $\mS' \otimes \mR$ as a metric of the
``block-diagonality'' of the matrix $\mS'$, where $\otimes$ indicates
element-wise multiplication and summation over all entries,
\ie $\mA \otimes \mB = \sum_{i,j} \emA_{i,j} \emB_{i,j}$.
Therefore, we can give the optimal connectivity pattern
between the adjacent layers by optimizing the following:
\begin{equation}
    \begin{split}
        &\min_{\mP, \mQ}~\mP \mS \mQ \otimes \mR \\
        &\mbox{s.t.}~\mP \in \mathcal{P}^{C^{\text{out}}}~\mbox{and}~~\mQ \in \mathcal{P}^{C^{\text{in}}}.
    \end{split}
    \label{eqn:optimization}
\end{equation}

However, the optimization over the set of permutation matrices is a non-convex and NP-hard problem,
which requires combinatorial search.
%
% Typically, combinatorial search is employed to solve this problem, which is computationally inefficient.
% \xth{Better to give a reference to a combinatorial search paper.}
%
To overcome the difficulty, we relax the feasible space to its convex hull
and alternatively optimize $\mP$ and $\mQ$. 
The Birkhoff-von Neumann theorem \citep{birkhoff1946three} states that the convex hull of
the set of permutation matrices is the set of doubly-stochastic
matrices\footnote{Doubly-stochastic matrices are non-negative square
matrices whose rows and columns sum to one.}, known as the
\textit{Birkhoff polytope}:
\begin{equation}
    \mathcal{B}^N = \{ \mX \in \R_{+}^{N \times N} : \mX \bm{1}_N =
    \bm{1}_N,~\mX^{\top} \bm{1}_N = \bm{1}_N \},
\end{equation}
where $\bm{1}_N$ denotes the column vector composed of $N$ ones.
% 
% We solve \cref{eqn:optimization} with \textit{coordinate descent}.
%
Then, we alternatively optimize $\mP$ and $\mQ$ on the Birkhoff polytope until convergence.
%
% When updating one variable, we consider the other as fixed.
%
For example, when optimizing $\mP$, the objective can be
formulated as follows:
\begin{equation}
    \begin{split}
        &\min_{\mP}~\mP \otimes \mR \mQ^{\top} \mS^{\top} \\
        &\mbox{s.t.}~\mP \in \mathcal{B}^{C^{\text{out}}}.
    \end{split}
    \label{eqn:lp}
\end{equation}
Similarly, we can have the objective for optimizing $\mQ$.

Note that the Birkhoff polytope is a \textit{simplex}. Therefore, the linear programming problem in \cref{eqn:lp} can be solved by efficient linear programming algorithms such as the 
network simplex method \citep{bonneel2011displacement}.
In addition, the theory of linear programming guarantees that at least
one of the solutions is achieved at the vertex of the simplex,
and the vertices of the Birkhoff polytope are precisely
the permutation matrices \citep{birkhoff1946three}.
Thus the solution to \cref{eqn:lp} is a naturally a permutation matrix without the need of any post-processing operation.
Furthermore, \cref{eqn:lp} can be viewed equivalently as the
\textit{optimal transport} problem, where an established computation
library\footnote{\url{https://github.com/rflamary/POT/}}
is available for efficient linear programming.

\subsection{Structured Regularization}
\label{sec:sparsification}
Permutation alone does not suffice to induce structurally sparse
convolutional weights, and 
we still need to impose a sparsity regularization to achieve
the desired sparsity structure.
Inspired by the sparsity-inducing penalty in \citet{liu2017learning},
we impose the structured $L_1$ regularization on
the permuted weight norm $\mS' = \mP \mS \mQ$.
Since the cardinality should be a power of 2,
suppose $\text{cardinality} = 2^{g-1}$ and here $g$ is defined as the group level,
as shown in \cref{fig:group-level}.
Given the group level $g$, the structured $L_1$
regularization is formulated as\footnote{For simplicity, we here compute the regularization of a single convolutional layer.
In the experiments, the
regularization is the summation of those of all the convolution layers.}
$\mathcal{L}_{\text{reg}} = \mS' \otimes \mR_g$,
where $\mR_g$ denotes the structured regularization matrix
as illustrated in \cref{fig:group-level}(b).
Intuitively, the proposed structured regularization aims to zero out the non-diagonal entries so as to make the permuted weight norm matrix $\mS'$ as much sparsified as possible. 
Furthermore, the regularization coefficient decays exponentially
as the group level grows as we desire balanced cardinality
distribution among the network.
In the end, the overall loss becomes
\begin{equation}
    \mathcal{L} = \mathcal{L}_{\text{data}} + \lambda \mathcal{L}_{\text{reg}},
    \label{eq:total-loss}
\end{equation}
where $\mathcal{L}_{\text{data}}$ denotes the regular
data loss (standard classification loss in the following experiments) and $\lambda$ is the balancing scalar coefficient.

\subsection{Criteria of Learning Cardinality}
\label{sec:criteria}

As the sparsity changes during the training process, we need to  determine the cardinality based on the structurally sparsified convolutional weights.
Since the weight norms corresponding to the valid connections constitute at least a certain proportion of the total weight norms,
we set a threshold $p$ to determine the group level $g$. 
Motivated by \cite{zhao2020dependency}, the group level $g$ can be determined by
\begin{equation}
    g = \max \{ g : \mS' \otimes \mU_g \ge p \sum_{i,j} \emS_{i,j},~g=1,2,\cdots \},
    \label{eq:group-level}
\end{equation}
where $p$ is a threshold set to 0.9 in all of the experiments
and $\mU_g$ is the relationship matrix \citep{zhang2019differentiable}
as illustrated in \cref{fig:group-level}(c) and specifies the valid neuron connections at group level $g$.
% 
% Please refer to the supplementary materials for the full
% structured sparsification algorithm.

\subsection{Implementation Details}
\label{sec:details}

\paragraph{Overall Configuration.}
Our implementation is based on the PyTorch library \citep{steiner2019pytorch}.
%\citep{steiner2019pytorch} library.
%
The proposed method is applied to the ResNet \citep{he2016deep} and DenseNet \citep{huang2017densely} families, and evaluated on the CIFAR-10 \citep{krizhevsky2009learning} and ImageNet \citep{ILSVRC15} datasets.
For the CIFAR-10 dataset, we follow the common practice of
data augmentation \citep{he2016deep,liu2017learning,xie2017aggregated}: zero-padding of 4 pixels on each side of the image, random crop of a $32 \times 32$ patch, and random horizontal flip.
For fair comparisons, we utilize the same network architecture as
\citet{liu2017learning}, and the model is trained on a single
GPU with a batch size of 64.
For the ImageNet dataset, we adopt the standard data augmentation
strategy \citep{vgg,he2016deep,xie2017aggregated}:
image resize such that the shortest edge is of 256 pixels,
random crop of a $224 \times 224$ patch,
and random horizontal flip.
The overall batch size is 256, which is distributed to 4 GPUs.
For both datasets, we employ the SGD optimizer with momentum 0.9.
%
%MH: add this line
% The source code and trained models will be made available
% to the public upon acceptance.

\begin{algorithm}[!t]
\caption{Training Pipeline.}\label{alg:training}
Initially update the permutation matrices $\mP$ and $\mQ$.\\
\For{$t := 1$ \textrm{to \#epochs}}
{
  Train for 1 epoch with the structured regularization; \\
  Solve \cref{eqn:optimization} in the main text to update the matrices $\mP$ and $\mQ$; \\
  Determine the current group levels $g$ by \cref{eq:group-level}; \\
  Update the structured regularization matrices; \\
  Adjust the coefficient $\lambda$.
}
\end{algorithm}

\paragraph{Training Protocol.}
For the first stage, we train a large model from scratch with the
structured regularization as described in Sec. 3.3.
At the end of each epoch,
we update the permutation matrices as in Sec. 3.2,
determine the current group levels as in Sec. 3.4,
adjust the structured regularization matrices accordingly,
and search for the coefficient $\lambda$ to meet the desired
compression rate as shown in \cref{appendix:penalty-adjustment}.
We train with a fixed learning rate of 0.1 for 100 epochs on the CIFAR-10 dataset
and 60 epochs on the ImageNet dataset and exclude
the weight decay due to the existence of the structured regularization.
The training pipeline is summarized in \cref{alg:training}.

\paragraph{Finetune Protocol.}
The remaining parameters are restored from the training stage and
the compressed model is finetuned with an initial learning rate of 0.1.
We finetune for 160 epochs on the CIFAR-10 dataset and the learning rate 
decays by a factor of 10 at 50\% and 75\% of the total epochs.
On the ImageNet dataset, the learning rate is decayed according to
the cosine annealing strategy \citep{loshchilov-ICLR17SGDR}
within 120 epochs.
For both datasets, a standard weight decay of $10^{-4}$ is adopted
to prevent overfitting.

\section{Experiments and Analysis}
In this section, we present the experimental results on
the CIFAR-10 and ImageNet datasets.
In addition, we carry out ablation studies to demonstrate 
the effectiveness of components of the proposed method. 
%our pipeline.
%
Please refer to \cref{appendix:details} for more detailed experimental results,
and our source codes are available at \url{https://github.com/Sakura03/StrucSpars}.

\begin{table}[!t]
    \renewcommand{\arraystretch}{1.25}
    \centering
    \caption{Network compression results on the CIFAR-10 \citep{krizhevsky2009learning} dataset.
    ``Baseline'' means the network without compression.
    The percentages in our method indicate the compression rate
    (measured by the reduction of ``\#Params.''), while those
    in other methods indicate the pruning ratio.
    }
    \vspace{1.2mm}
    \resizebox{0.475\textwidth}{!}{
        \begin{tabular}{lllr}
            \toprule[1.5pt]
            Methods   & \#Params.($10^5$) $\downarrow$ & FLOPs ($10^7$) $\downarrow$ & Acc.(\%) $\uparrow$\\
            \hline\hline
            \multicolumn{4}{c}{\textbf{ResNet-20}}\\
            Baseline & 2.20 & 3.53 & 91.70 ($\pm$0.12) \\
            \cdashline{1-4}[2pt/3pt]
            Slimming-40\% & 1.91 ($\pm$0.00) & {\bf 3.10} ($\pm$0.02) & 91.74 ($\pm$0.35) \\
            StrucSpars-20\%     & {\bf 1.76} ($\pm$0.00) & 3.18 ($\pm$0.07) & {\bf 91.79} ($\pm$0.23) \\
            \cdashline{1-4}[2pt/3pt]
            Slimming-60\% & 1.36 ($\pm$0.02) & {\bf 2.24} ($\pm$0.01) & 89.68 ($\pm$0.38) \\
            StrucSpars-40\%     & {\bf 1.31} ($\pm$0.01) & 2.58 ($\pm$0.00) & {\bf 91.42} ($\pm$0.04) \\
            \hline\hline
            \multicolumn{4}{c}{\textbf{ResNet-56}}\\
            Baseline & 5.90 & 9.16 & 93.50 ($\pm$0.19)\\
            \cdashline{1-4}[2pt/3pt]
            Slimming-60\% & 4.15 ($\pm$0.03) & 5.75 ($\pm$0.10) & 93.10 ($\pm$0.25) \\
            StrucSpars-30\%     & 4.08 ($\pm$0.05) & 7.17 ($\pm$0.20) & {\bf 94.19} ($\pm$0.16) \\
            StrucSpars-50\%     & {\bf 2.96} ($\pm$0.03) & {\bf 4.81} ($\pm$0.03) & 93.70 ($\pm$0.06) \\
            \cdashline{1-4}[2pt/3pt]
            Slimming-80\% & 2.33 ($\pm$0.04) & {\bf 3.50} ($\pm$0.02) & 91.01 ($\pm$0.02) \\
            StrucSpars-60\%     & 2.34 ($\pm$0.08) & 4.20 ($\pm$0.08) & {\bf 93.48} ($\pm$0.13) \\
            StrucSpars-70\%     & {\bf 1.80} ($\pm$0.00) & 3.52 ($\pm$0.16) & 93.25 ($\pm$0.02) \\
            \hline\hline
            \multicolumn{4}{c}{\textbf{ResNet-110}}\\
            Baseline & 11.47 & 17.59 & 94.62 ($\pm$0.22) \\\cdashline{1-4}[2pt/3pt]
            Slimming-40\% & 9.24 ($\pm$0.03) & 12.55 ($\pm$0.00) & 94.49 ($\pm$0.12) \\
            StrucSpars-20\%     & 9.12 ($\pm$0.06) & 14.76 ($\pm$0.02) & {\bf 94.78} ($\pm$0.11) \\
            StrucSpars-40\%     & {\bf 6.69} ($\pm$0.24) & {\bf 11.60} ($\pm$0.01) & 94.55 ($\pm$0.18)  \\
            \cdashline{1-4}[2pt/3pt]
            Slimming-60\% & 8.15 ($\pm$0.03) & {\bf 10.66} ($\pm$0.00) & 94.29 ($\pm$0.11) \\
            StrucSpars-30\%     & 7.89 ($\pm$0.03) & 12.47 ($\pm$0.01) & {\bf 94.69} ($\pm$0.08) \\
            StrucSpars-60\%     & {\bf 5.41} ($\pm$0.02) & {\bf 10.66} ($\pm$0.01) & 94.42 ($\pm$0.04)  \\
            \toprule[1.5pt]
        \end{tabular}
    }\label{tab:cifar}
\end{table}

\subsection{Results on CIFAR-10}
We first compare our proposed method with the Network Slimming \citep{liu2017learning} approach
on the CIFAR-10 dataset.
The Network Slimming approach is a representative filter pruning method that
compresses CNNs by pruning less important filters.
%
% We use the Resnet~\cite{he2016deep} architecture with various
% depths and compression rates.
%
As the experimental results on the CIFAR-10 dataset are somewhat random, we repeat the train-compress-finetune pipeline for 10 times and record the mean and standard deviation (std).
As shown in \cref{tab:cifar}, the proposed structured sparsification performs favorably under various compression rates.
%remarkable performance
%
For ResNet-110, with 60\% parameters compressed,
the structured sparsification can still achieve a 94.42\% top-1 accuracy which is nearly equal
to the performance of the baseline method without compression. 
Compared with the Network Slimming, our method consistently performs better,
especially under high compression rates.
Experiments on the CIFAR-10 dataset demonstrate that our method is able to compress
CNNs with negligible performance drop and favorable accuracy against pruning methods
such as Network Slimming.

\begin{table}[!t]
    \centering
    \caption{Network compression results on the ImageNet \citep{ILSVRC15} dataset.
      The center-crop validation accuracy is reported. ``Baseline'' means the network without compression.
      The percentages in the table have the same meaning as those in \cref{tab:cifar}.
    }
    \vspace{1.2mm}
    \resizebox{0.475\textwidth}{!}{
        \begin{tabular}{lccr}
            \toprule[1.5pt]
            Methods & \#Params.($10^6$) $\downarrow$ & GFLOPs $\downarrow$ & Acc.(\%) $\uparrow$\\
            \hline\hline
            \multicolumn{4}{c}{\textbf{ResNet-50}}\\
            Baseline & \makebox[2em][r]{25.6} & \makebox[2em][r]{4.14} & 77.10 \\
            \cdashline{1-4}[2pt/3pt]
            NISP-A & \makebox[2em][r]{18.6} & \makebox[2em][r]{$\approx$2.97} & 72.75 \\
            Slimming-20\% & \makebox[2em][r]{17.8} & \makebox[2em][r]{2.81} & 75.12 \\
            Taylor-19\% & \makebox[2em][r]{17.9} & \makebox[2em][r]{2.66} & 75.48 \\
            FPGM-30\% & \makebox[2em][r]{N/A} & \makebox[2em][r]{\textbf{2.39}} & 75.59 \\
            StrucSpars-35\% & \makebox[2em][r]{\textbf{17.2}} & \makebox[2em][r]{3.12} & \textbf{76.82} \\
            \cdashline{1-4}[2pt/3pt]
            ThiNet-30\% & \makebox[2em][r]{16.9} & \makebox[2em][r]{$\approx$2.62} & 72.04 \\
            NISP-B & \makebox[2em][r]{14.3} & \makebox[2em][r]{$\approx$2.29} & 72.07 \\
            ABCPrunner-80\% & \makebox[2em][r]{11.8} & \makebox[2em][r]{1.89} & 73.86 \\
            Taylor-28\% & \makebox[2em][r]{14.2} & \makebox[2em][r]{2.25} & 74.50 \\
            DSA-50\% & \makebox[2em][r]{N/A} & \makebox[2em][r]{2.07} & 74.69 \\
            Hinge-46\% & \makebox[2em][r]{N/A} & \makebox[2em][r]{$\approx$1.93} & 74.70 \\
            FPGM-40\% & \makebox[2em][r]{N/A} & \makebox[2em][r]{1.93} & 74.83 \\
            HRank-36\% & \makebox[2em][r]{16.2} & \makebox[2em][r]{2.30} & 74.98 \\
            StrucSpars-65\% & \makebox[2em][r]{\textbf{10.3}} & \makebox[2em][r]{\textbf{1.67}} & \textbf{75.10} \\
            \cdashline{1-4}[2pt/3pt]
            ThiNet-50\% & \makebox[2em][r]{12.4} & \makebox[2em][r]{$\approx$1.83} & 71.01 \\
            Taylor-44\% & \makebox[2em][r]{7.9} & \makebox[2em][r]{1.34} & 71.69 \\
            HRank-46\% & \makebox[2em][r]{13.8} & \makebox[2em][r]{1.55} & 71.98 \\
            Slimming-50\% & \makebox[2em][r]{11.1} & \makebox[2em][r]{1.87} & 71.99 \\
            StrucSpars-85\% & \makebox[2em][r]{\textbf{5.6}} & \makebox[2em][r]{\textbf{0.90}} & \textbf{72.47} \\
            \hline\hline
            \multicolumn{4}{c}{\textbf{ResNet-101}}\\
            Baseline & \makebox[2em][r]{44.5} & \makebox[2em][r]{7.87} & 78.64 \\\cdashline{1-4}[2pt/3pt]
            FPGM-30\% & \makebox[2em][r]{N/A} & \makebox[2em][r]{\textbf{4.55}} & 77.32 \\
            Taylor-25\% & \makebox[2em][r]{31.2} & \makebox[2em][r]{4.70} & 77.35 \\
            StrucSpars-40\% & \makebox[2em][r]{\textbf{26.7}} & \makebox[2em][r]{5.05} & \textbf{78.16} \\
            \cdashline{1-4}[2pt/3pt]
            BN-ISTA-v1 & \makebox[2em][r]{17.3} & \makebox[2em][r]{3.69} & 74.56 \\
            BN-ISTA-v2 & \makebox[2em][r]{23.6} & \makebox[2em][r]{4.47} & 75.27 \\
            ABCPrunner-80\% & \makebox[2em][r]{17.7} & \makebox[2em][r]{3.16} & 75.82 \\
            Taylor-45\% & \makebox[2em][r]{20.7} & \makebox[2em][r]{\textbf{2.85}} & 75.95 \\
            Slimming-50\% & \makebox[2em][r]{20.9} & \makebox[2em][r]{3.16} & 75.97 \\
            SCOP-B & \makebox[2em][r]{18.8} & \makebox[2em][r]{3.13} & 77.36 \\
            DMC-56\% & \makebox[2em][r]{N/A} & \makebox[2em][r]{3.46} & 77.40 \\
            StrucSpars-65\% & \makebox[2em][r]{\textbf{16.5}} & \makebox[2em][r]{2.98} & \textbf{77.62} \\
            \cdashline{1-4}[2pt/3pt]
            Taylor-60\% & \makebox[2em][r]{13.6} & \makebox[2em][r]{1.76} & 74.16 \\
            ABCPrunner-50\% & \makebox[2em][r]{12.9} & \makebox[2em][r]{1.98} & 74.76 \\
            StrucSpars-80\% & \makebox[2em][r]{\textbf{10.6}} & \makebox[2em][r]{\textbf{1.70}} & \textbf{75.73} \\
            \hline\hline
            \multicolumn{4}{c}{\textbf{DenseNet-201}}\\
            Baseline & \makebox[2em][r]{20.0} & \makebox[2em][r]{4.39} & 77.88 \\\cdashline{1-4}[2pt/3pt]
            Taylor-40\% & \makebox[2em][r]{\textbf{12.5}} & \makebox[2em][r]{\textbf{3.02}} & 76.51 \\
            StrucSpars-38\% & \makebox[2em][r]{13.1} & \makebox[2em][r]{3.53} & \textbf{77.43} \\
            \cdashline{1-4}[2pt/3pt]
            Taylor-64\% & \makebox[2em][r]{\textbf{9.0}} & \makebox[2em][r]{2.21} & 75.28 \\
            StrucSpars-60\% & \makebox[2em][r]{9.2} & \makebox[2em][r]{\textbf{2.10}} & \textbf{75.86} \\
            \cdashline{1-4}[2pt/3pt]
            \toprule[1.5pt]
        \end{tabular}
    }\label{tab:compression}
\end{table}

\subsection{Results on ImageNet}
\cref{tab:compression} shows the evaluation results of the proposed method against the recent representative network pruning approaches,
including ThiNet \citep{luo2017thinet}, Slimming \citep{liu2017learning},
NISP \citep{yu2018nisp}, BN-ISTA \citep{ye2018rethinking}, FPGM
\citep{he2019filter}, Taylor \citep{molchanov2019importance},
ABCPrunner \citep{ijcai2020-94}, HRank \citep{lin2020hrank}, Hinge \citep{li2020group},
DMC \citep{Gao_2020_CVPR}, DSA \citep{ning2020dsa}, and SCOP \citep{tang2020scop}.
%
%MH: tone down 
%The statistics clearly indicate that our approach outperforms the previous
%pruning-based counterparts by a large margin.
%
Overall, the structured sparsification method performs favorably against the previous
network compression methods under different settings. 
%
%MH: do not argue.... try not use unnecessary adjective
%We argue the reason lies in the fact that discarding the entire filters will dramatically degrade the representational capacity of the network, especially when the pruning ratio is high.
These performance gains achieved by our method can be attributed to the fact that 
discarding the entire filters will negatively affect the representational strength of the network model, especially when the pruning ratio is high, \eg 50\%.
In contrast, our method removes only a proportion of neuron connections and preserves all of the filters, thereby making a mild impact on the model capacity.
In addition, it is known that pruning neuron connections would eliminate the information flow and affect performance.
To alleviate this issue, the learnable channel shuffle mechanism assists the information exchange among different groups, thereby minimizing the potential negative impact. 

\subsection{Ablation Studies}

\paragraph{Accuracy \vs~Complexity.}

\begin{figure}[!t]
    %\resizebox{0.60\textwidth}{!}{%%
    \centering
    \begin{overpic}[width=\linewidth]{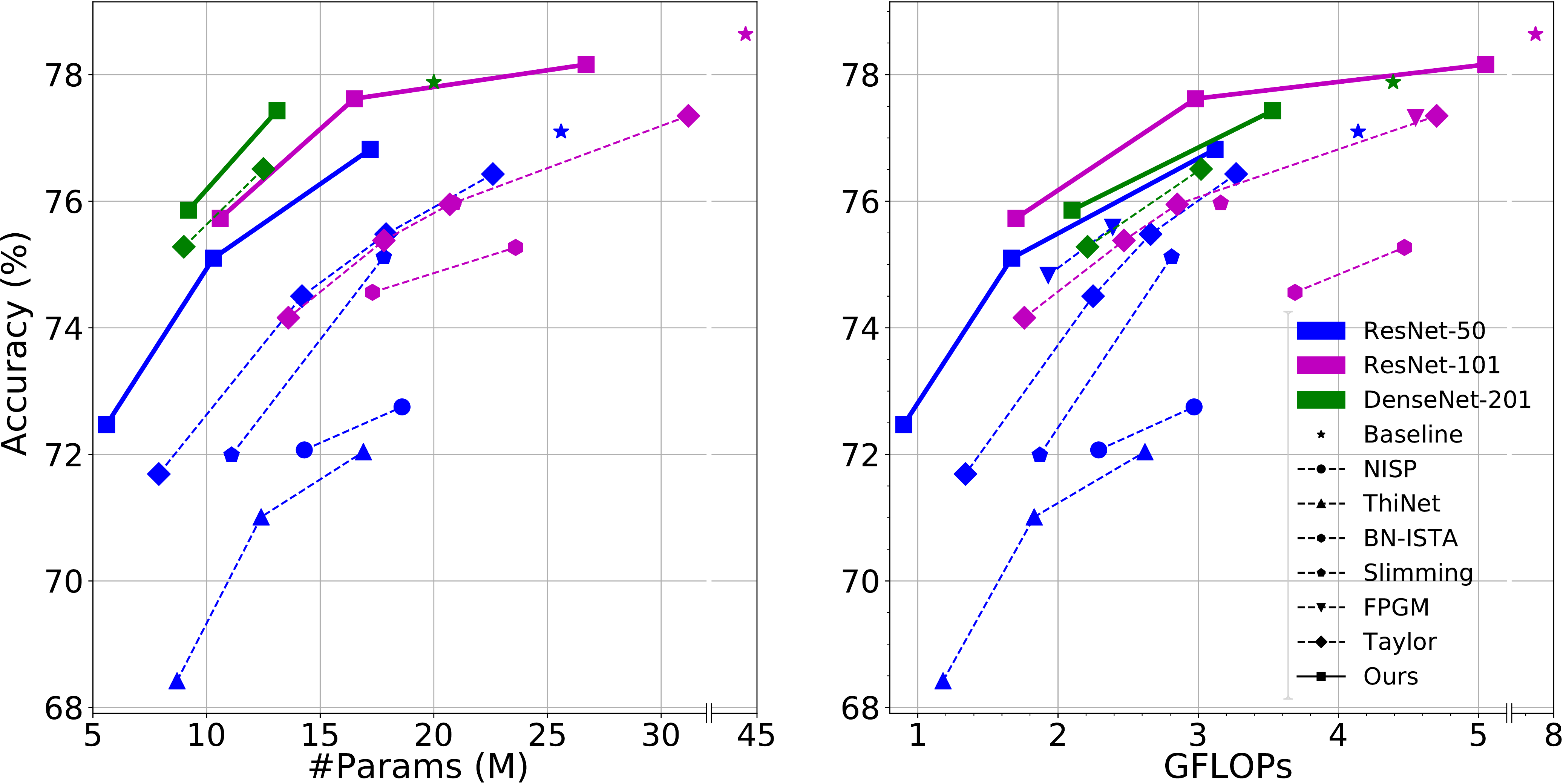}
    \end{overpic}
    \caption{Accuracy-complexity Trade-off on the ImageNet~\citep{ILSVRC15} dataset.
    (Upper left is better.)
    }
    \label{fig:trade-off}
\end{figure}

As shown in \cref{fig:trade-off}, the proposed structured sparsification is designed to make sound accuracy-complexity trade-off.
On the ImageNet \citep{ILSVRC15} dataset,
a slight top-1 accuracy drop of 0.28\% is compromised
for about 25\% complexity reduction on the ResNet-50 backbone,
and an accuracy loss of 1.02\% for about 60\% reduction on ResNet-101.
Furthermore, high compression rates can be achieved in our methodology while maintaining competitive performance.
%
%MH: do not use useless adjective "remarkably". Tone it down
%Remarkably, our method achieves an accuracy of 72.47\% with only about $20\% complexity of the ResNet-50 backbone, which even outperforms the pruning methods with $2\times$ complexity.
%
It is worth noticing that our method achieves an accuracy of 72.47\% with only about 20\% complexity of the ResNet-50 backbone,
which performs favorably against the pruning methods with two times complexity.

\paragraph{Learned Channel Shuffle Mechanism.}
We evaluate the effectiveness of our learned channel shuffle
mechanism on the ResNet backbone with a compression rate of 65\%.
We use the following five settings for performance evaluation:
\begin{enumerate}[itemsep=0.1mm,leftmargin=5mm]
%MH: check this sentence... which sparsitication? simple sparticfication or structured sparsificaiton?
    \item \textsc{Finetune}:
    The preserved parameters after compression are restored
    and the compressed model is finetuned.
    For the other four settings, the parameters of the compressed model are re-initialized for the finetune stage.
    
    \item \textsc{FromScratch}:
    We keep the learned channel connectivity, \ie $\mP$ and $\mQ$, from the training stage, and train the model from randomly re-initialized weights.
    
    \item \textsc{ShuffleNet}:
    The same channel shuffle operation in the ShuffleNet
    \citep{zhang2018shufflenet} is adopted.
    Specifically, if a convolution is of cardinality $G$ and
    has $G \times N$ output channels, then the channel shuffle operation
    is equivalent to reshaping the output channel dimension
    into $(G, N)$, transposing and flattening it back. 
    Compared with \textsc{ShuffleNet}, the way of channel shuffle
    is learned rather than pre-defined in our method,
    \ie \textsc{Finetune} and \textsc{FromScratch}.
    
    \item \textsc{Random}:
    The permutation matrices $\mP$ and $\mQ$ are randomly given,
    independent of the training process.
    
    \item \textsc{NoShuffle}:
    The channel shuffle operations are removed,
    \ie $\mP$ and $\mQ$ are identity matrices.
\end{enumerate}
\begin{table}[!b]
    \centering
    \caption{Ablation study of different channel shuffle operations on the ImageNet dataset~\citep{ILSVRC15}.
    }
    \vspace{2mm}
    \resizebox{\linewidth}{!}{
        \begin{tabular}{lcccc}
            \toprule[1.5pt]
            Config. & \multicolumn{2}{c}{\textbf{ResNet-50}-65\%} &
            \multicolumn{2}{c}{\textbf{ResNet-101}-65\%} \\
            \midrule
            Acc. & Top-1 & Top-5 & Top-1 & Top-5 \\
            \midrule
            \textsc{Finetune} & 75.10 & 92.52 & 77.62 & 93.72 \\
            \textsc{FromScratch} & 75.02 & 92.46 & 77.14 & 93.53 \\
            \textsc{ShuffleNet} & 74.97 & 92.41 & 76.91 & 93.38 \\
            \textsc{Random} & 69.45 & 89.45 & 73.16 & 91.44 \\
            \textsc{NoShuffle} & 73.30 & 91.39 & 75.31 & 92.64 \\ 
            \toprule[1.5pt]
        \end{tabular}
    }\label{tab:channel}
\end{table}

\begin{table}[!t]
  \centering
  \caption{
    Wall-time acceleration of the structured sparsification method.
    }
   \resizebox{\linewidth}{!}{
      \begin{tabular}{lcccccccc}
          \toprule[1.5pt]
          Model & GFLOPs & Avg. Runtime (ms) & FPS \\
          \midrule
          ResNet-50  & 4.14 & 80.2 & 12.4 \\
          \name-35\% & 3.12 & 68.2 & 14.7 \\
          \name-65\% & 1.67 & 61.3 & 16.3 \\
          \name-85\% & 0.90 & 53.5 & 18.7 \\
          \toprule[1.5pt]
      \end{tabular}
   }
   \vspace{-3pt}
  \label{tab:wall-time}
\end{table}

The results are demonstrated in \cref{tab:channel}.
%
% First, we observe the compressed models trained from scratch give
% comparable performance with those finetuned from the first stage,
% indicating the restored parameters are not crucial to the final performance
% and that the searched network architecture is of superiority.
%
%MH: check this sentence
First, the finetuned models perform slightly better than 
those trained from scratch,
which implies that the preserved parameters
take an essential role in the final performance.
%
%MH: check this sentence
Furthermore, the model with learned channel shuffle mechanism, \ie neuron connectivity, performs the best among all settings.
The channel shuffle mechanism in the ShuffleNet~\citep{zhang2018shufflenet}
is effective as it outperforms the no-shuffle counterpart.
However, it can be further improved by a learning-based strategy.
Interestingly, the random channel shuffle scheme performs the worst, even worse than the no-shuffle scheme.
This implies that learning the channel shuffle operation is a challenging task, and randomly gathering channels from different groups gives no benefits.

\paragraph{Wall-time Acceleration}

We measure the wall-time of the ResNet-50 backbone and the compressed variants
on a single core of the Intel E5-2603 v4@1.70GHz CPU.
As in \cref{tab:wall-time}, we report the average runtime and the frames per second (FPS) of different models when processing the $224 \times 224$ images.
It can be seen that our method can result in actual wall-time acceleration in the real-world scenarios.

\section{Future Work}

We discuss three potential directions for future work
along the line of our work.
\begin{enumerate}[label=(\roman*)]
    \item {Data-Driven Structured Sparsification.}
    In this work, the structured regularization is uniformly imposed on the convolutional weights, thus the learned cardinality distribution is prone to uniformity.
    Besides, the structured regularization is calculated independently of the data loss (see \cref{eq:total-loss}),
    so the gradients of the structured regularization are not guided by the task information,
    thus leading to task-agnostic cardinalities of the compressed models.
    Nevertheless, it is possible to obtain model structures with better performance if the structured regularization could be guided by the back-propagated signals of the data loss.
    The optimization-based meta-learning techniques \citep{pmlr-v70-finn17a} can be exploited
    for this purpose.
    % Currently, the gradients of the data loss and those of
    % the structured regularization are computed independently
    % in each training iteration (see \cref{eq:total-loss}).
    % %
    % Thus, the structured regularization is imposed uniformly on the
    % convolutional layers, and the learned cardinality distribution is
    % task-agnostic and prone to uniformity.
    % %
    % However, better cardinality distribution may be achieved 
    % if the structured sparsification is guided
    % by the back-propagated signals of the data loss.
    % %
    % Thus, optimization-based meta-learning techniques
    % \cite{pmlr-v70-finn17a} can be exploited for this purpose.

    \item {Progressive Sparsification Solution.}
    Typically, finetune-free compression techniques are desired in
    practical applications \citep{cheng2018recent}.
    Therefore, the sparsified weights can be removed
    progressively during training, and the architecture
    search as well as model training can be completed in a single training pass.
    
    \item {Combination with Filter Pruning Techniques.}
    As the entire feature maps are reserved in our approach,
    the reduction of memory footprint is limited.
    This issue can be addressed by combining with the filter pruning techniques,
    which is non-trivial as uniform filter pruning is required within each group.
    It is of great interest to exploit group sparsity constraints \citep{yoon2017combined}
    to achieve such uniform sparsity.
    % The proposed structured sparsification reduces the number of model parameters
    % by structurally sparsifying the convolutional weights and
    % converting vanilla convolutions into group convolutions.
    % %
    % However, the number of model parameters could be further reduced
    % by uniformly pruning filters within each group as a post-processing step,
    % where group sparsity constraints~\cite{yoon2017combined} could be enforced
    % in order to achieve uniform sparsity.
    % % We argue that the combination is non-trivial because
    % the structure of sparsified weights may be impaired by filter
    % pruning.
\end{enumerate}

\section{Conclusion}

In this work, we propose a structured sparsification method for efficient network compression,
where the structurally sparse representations of the convolutional weights are induced
and the inter-group information flow is facilitated by the learnable channel shuffle.
The compressed model can be readily incorporated in modern deep learning frameworks
thanks to their support for the group convolution.
%
% The problem of inter-group communication is further solved via
% the learnable channel shuffle mechanism.
%
The proposed approach is flexible with special network structures and
highly compressible with negligible performance degradation,
as validated on the CIFAR-10 and ImageNet datasets.
%
% We suggest that future research directions include extending the structured sparsification as a data-driven method and combining with filter pruning techniques, for which detailed discussions are in the supplementary materials.

%
%MH: I do not know what you want to say here
%Despite the existence of potential limitations and improvements,
%the feasibility of our approach is demonstrated in this work,
%indicating a new path is paved for network compression and that
%techniques of structured sparsification can be a fruitful
%future research direction.
%MH: check this sentence
% The proposed method performs favorably against the state-of-the-art
% model compression approaches even with compression rate.
%The favorable performance of the proposed method against
%the state-of-the-art model compression approaches indicates
%a new path is paved for efficient network compression and that
%techniques of structured sparsification can be a fruitful
%future research direction.
%
% In addition, experimental evaluation against the state-of-the-art
% compression approaches shows techniques of structured sparsification
% can be a fruitful future research direction.

\bibliography{uai2021-template}

\clearpage

\appendix
\onecolumn

\section{Structured Regularization in General Form} \label{appendix:reg}
Generally, we can relax the constraints that both $C^{\text{in}}$
and $C^{\text{out}}$ are powers of 2, and assume both
$C^{\text{in}}$ and $C^{\text{out}}$ have many factors of 2.
Under this assumption, the potential candidates of cardinality
are still restricted to powers of 2.
Specifically, if the \textit{greatest common divisor} of
$C^{\text{in}}$ and $C^{\text{out}}$ can be factored as
\begin{equation}
    \textrm{gcd}(C^{\text{in}}, C^{\text{out}}) = r = 2^u \cdot z,
\end{equation}
where $z$ is an odd integer,
then the potential candidates of the group level $g$ are
$\{1, 2, \cdots, u+1\}$.
For example, if the minimal $u$ is 4 among all convolutional
layers\footnote{The standard DenseNet \citep{huang2017densely}
family satisfies this condition.},
the potential candidates of cardinality are $\{1, 2, 4, 8, 16\}$,
giving adequate flexibility of the compressed model.
The structured regularization and the relationship matrix
corresponding to each group level are designed in a similar way.
For clarity, we provide an exemplar implementation based on the NumPy library.

% \begin{figure*}
% \begin{minipage}[!b]{0.85\textwidth}
\lstdefinestyle{customc}{
  belowcaptionskip=1\baselineskip,
  breaklines=true,
  frame=L,
  xleftmargin=18.0pt,
  language=C,
  showstringspaces=false,
  basicstyle=\footnotesize\ttfamily,
  keywordstyle=\bfseries\color{green!40!black},
  commentstyle=\itshape\color{purple!40!black},
  identifierstyle=\color{blue},
  stringstyle=\color{orange},
  numbers=left,
}
\lstdefinestyle{customasm}{
  belowcaptionskip=1\baselineskip,
  frame=L,
  xleftmargin=\parindent,
  language=[x86masm]Assembler,
  basicstyle=\footnotesize\ttfamily,
  commentstyle=\itshape\color{purple!40!black},
}
\lstset{escapechar=@,style=customc}
\begin{lstlisting}[captionpos=b,language=Python]
import numpy as np

def struc_reg(dim1, dim2, level=None, power=0.5):
    r"""
    Compute the structured regularization matrix.
    
    Args::
        dim1 (int): number of output channels.
        dim2 (int): number of input channels.
        level (int or None): current group level.
                Specify 'None' if the cost matrix is desired.
        power (float): decay rate of the reg. coefficients.
    
    Return::
        Structured regularization matrix.
    """
    reg = np.zeros((dim1, dim2))
    assign_val(reg, 1., level, power)
    return reg

def assign_val(arr, val, level, power):
    dim1, dim2 = arr.shape
    if dim1 % 2 != 0 or dim2 % 2 != 0 or level == 0:
        return
    else:
        _l = None if level is None else level - 1
        arr[dim1//2:, :dim2//2] = val
        arr[:dim1//2, dim2//2:] = val
        assign_val(arr[dim1//2:, dim2//2:], val*power, _l, power)
        assign_val(arr[:dim1//2, :dim2//2], val*power, _l, power)
\end{lstlisting}
% \end{minipage}
% \end{figure*}

\section{Dynamic Penalty Adjustment}
\label{appendix:penalty-adjustment}
As the desired compression rate is customized according to user preference,
manually choosing an appropriate regularization coefficient $\lambda$ in \cref{eq:total-loss}
for each experimental setting is extremely inefficient.
To alleviate this issue, we dynamically adjust $\lambda$ based on the sparsification progress.
The algorithm is summarized in \cref{alg:penalty-adjustment}.

Concretely, after the $t^{th}$ training epoch, we first determine the current group level
$g_{t}$ of each convolutional layer according to \cref{eq:group-level}.
Then, we define the model sparsity based on the reduction of model parameters.
For the $l^{th}$ convolutional layer, the number of parameters is reduced by a factor
of $2^{g^{l}_{t}-1}$, where $2^{g^{l}_{t}-1}$ is the cardinality.
Thus, the original number of parameters and the reduced one are given by
\begin{equation}
    p^{l} = C^{l} \times C^{l+1} \times k^{l} \times k^{l},
    \quad
    \hat{p}^{l}_{t} = \frac{p^{l}}{2^{g^{l}_{t}-1}}.
    \label{eqn:orig-reduced-params}
\end{equation}
Here, $C^{l}$ and $k^{l}$ denote the input channel number and
the kernel size of the $l^{\rm th}$ convolutional layer.
Therefore, the current model sparsity is calculated as
\begin{equation}
    r_t = \frac{\sum_l \hat{p}^{l}_{t}}{\sum_l p^{l}}.
    \label{eqn:model-sparsity}
\end{equation}

Afterwards, we assume the model sparsity grows linearly,
and calculate the expected sparsity gain.
If the expected sparsity gain is not met, \ie
\begin{equation}
  r_{t}-r_{t-1}<\frac{r-r_{t-1}}{N-t+1},
\end{equation}
where $N$ is the
total training epoch number and $r$ is the target sparsity,
we increase $\lambda$ by $\Delta_{\lambda}$.
If the model sparsity exceeds the target, \ie $r_t>r$,
we decrease $\lambda$ by $\Delta_{\lambda}$.

In all experiments, the coefficient is initialized from $\lambda_1=0$ and
$\Delta_{\lambda}$ is set to $2 \times 10^{-6}$.

\begin{algorithm}[!t]
  \caption{Dynamically adjust $\lambda$}\label{alg:penalty-adjustment}
  Initialize $\lambda_1 = 0$, $r_0=0$, $N=\text{\#epochs}$, $r=\text{target sparsity}$\\
  \For( \textit{train for 1 epoch}){$t := 1$ to $N$}
  {
    Determine the current group levels $g$; \\
    Compute the current sparsity by \cref{eqn:orig-reduced-params,eqn:model-sparsity} \\
    \uIf{$r_t - r_{t-1} < \frac{r-r_{t-1}}{N-t+1}$}{
      $\lambda_{t+1} = \lambda_t + \Delta_{\lambda}$\
    }
    \uElseIf{$\mS_t > r$}{
      $\lambda_{t+1} = \lambda_t - \Delta_{\lambda}$\
    }
  }
\end{algorithm}

\begin{table*}[!b]
    \centering
    \caption{
    Performance along the timeline of our approach.
    The evaluation is performed on the ImageNet dataset.
    }
    \begin{tabular}{lcccccccc}
        \toprule[1.5pt]
        Backbone & \multicolumn{3}{c}{\textbf{ResNet-50}} &
        \multicolumn{3}{c}{\textbf{ResNet-101}} &
        \multicolumn{2}{c}{\textbf{DenseNet-201}} \\
        \midrule
        Compression Rate & 35\% & 65\% & 85\% & 40\% & 65\% & 80\% & 38\% & 60\% \\
        \midrule
        Pre-compression Acc.  & 69.07 & 66.36 & 64.30 & 69.56 & 67.13 & 64.20 & 69.10 & 66.26 \\
        Post-compression Acc. & 60.92 & 42.78 &  8.82 & 65.78 & 58.63 & 18.57 & 66.15 & 17.35 \\
        Finetune Acc.         & 76.82 & 75.10 & 72,47 & 78.16 & 77.62 & 75.73 & 77.43 & 75.86 \\
        threshold $p$         & 0.127 & 0.115 & 0.125 & 0.095 & 0.090 & 0.103 & 0.098 & 0.115 \\
        \toprule[1.5pt]
    \end{tabular}
    \label{tab:experimental-details}
\end{table*}

\section{Experimental Details} \label{appendix:details}
In this section, we provide more results and details of our experiments.
We provide the loss and accuracy curves along with the performance after
each stage in \cref{appendix-sec:curve},
and analyze the compressed model architectures in \cref{appendix-sec:arch}.

\subsection{Training Dynamics}\label{appendix-sec:curve}

\begin{figure*}[!t]
  \centering
  \begin{overpic}[width=\textwidth]{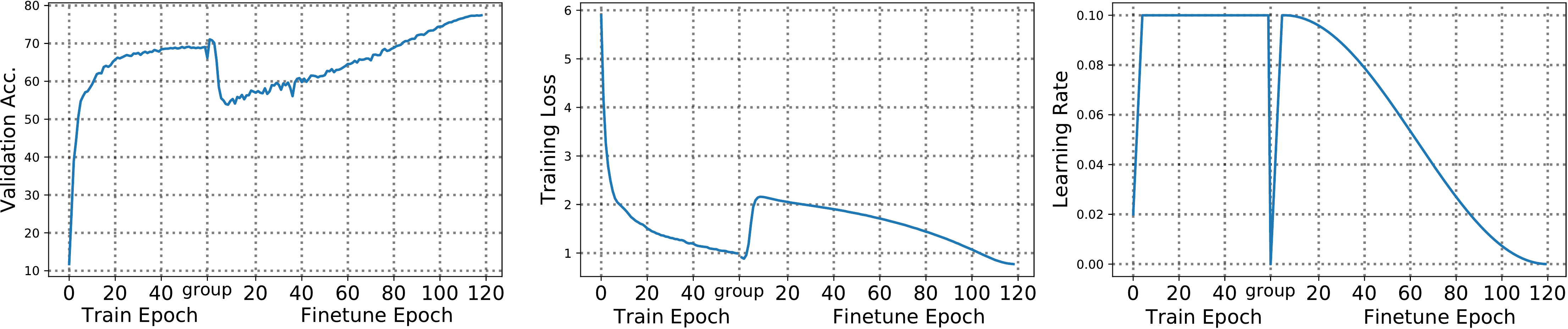}
  \end{overpic}
  \caption{
    Training dynamics of the full structured sparsification pipeline.
    We plot the training and finetune curves of the DenseNet-201
    backbone with a compression rate of 38\%.
    At the end of the 60$^{th}$ epoch of the training stage,
    we compress the network following our criteria.
    Then, we finetune for 120 epochs to recover performance.
  }\label{fig:training-curve}
\end{figure*}

We first provide the pre- and post-compression accuracy along with the finetune accuracy
of our pipeline in \cref{tab:experimental-details}.
During compression, we use a \textit{binary search} to decide the threshold $p$ of
the grouping criteria (\cref{eq:group-level}) so that the network can be compressed at the desired
compression rate.
The searched thresholds are also illustrated.
Apart from this, we further provide the training and finetune curves in \cref{fig:training-curve}.
In the training stage, the accuracy gradually increases till saturation,
and then the compression leads to a slight performance drop.
Finally, the performance is recovered in the finetune stage.

\subsection{Compressed Architectures}\label{appendix-sec:arch}
\begin{figure*}[!h]
  \centering
  \begin{overpic}[width=\textwidth]{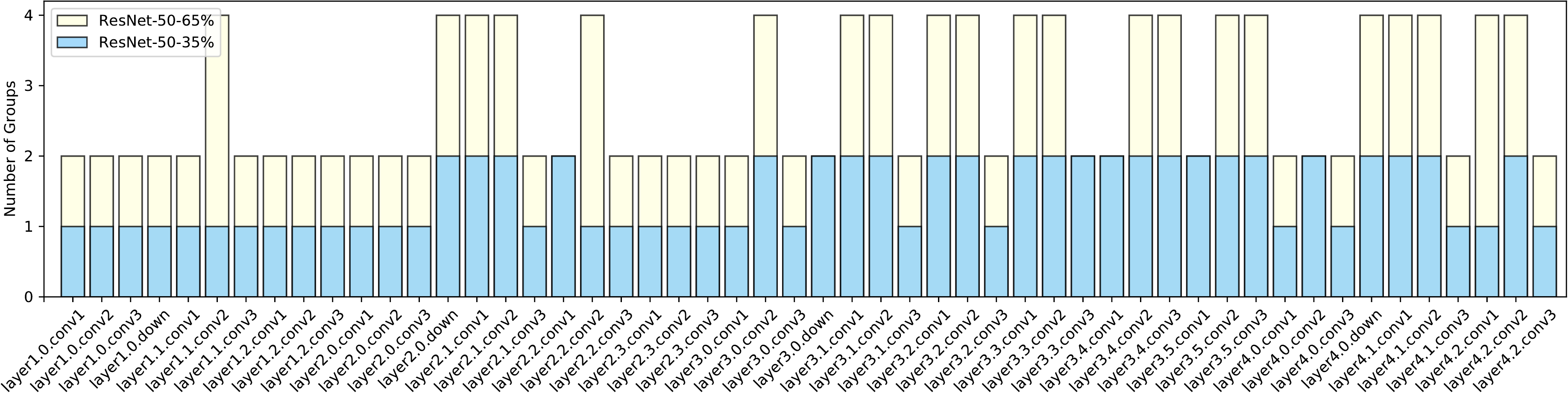}
  \end{overpic}
  \caption{
    Learned cardinalities of the ResNet-50 backbone with the compression rates
    of 35\% and 65\%.
  }
  \label{fig:cardinality-res50}
\end{figure*}

\begin{figure*}[!h]
  \centering
  \begin{overpic}[width=\textwidth]{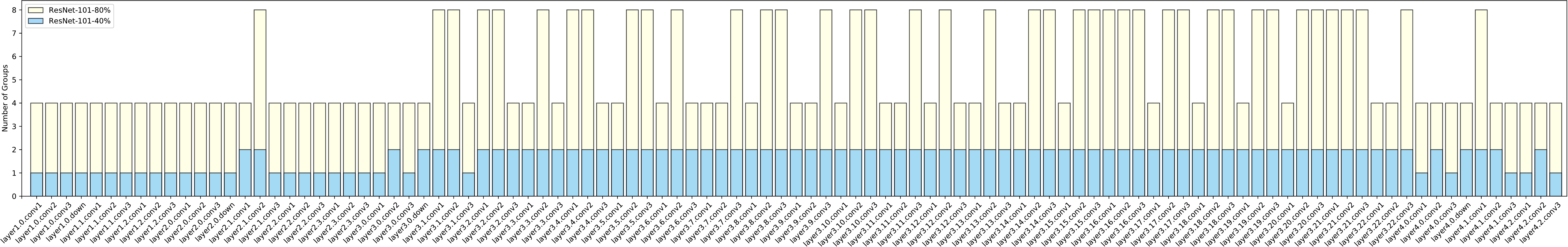}
  \end{overpic}
  \caption{
    Learned cardinalities of the ResNet-101 backbone with the compression rates
    of 40\% and 80\%.
  }
  \label{fig:cardinality-res101}
\end{figure*}

We illustrate the compressed architectures by showing the cardinality of each
convolution layer in \cref{fig:cardinality-res50} and \cref{fig:cardinality-res101}.
Note that our method is applied to all convolution operators, \ie both
$3 \times 3$ convolutions and $1 \times 1$ convolutions,
so a high compression rate, \eg 80\%, can be achieved.
As discussed in Sec. 4.4, the learned cardinality distribution is
prone to uniformity, but there are still certain patterns.
For example, shallow layers are relatively more difficult to be compressed.
A possible explanation is that shallow layers have fewer filters,
so a large cardinality will inevitably eliminate the communication
between certain groups.
Moreover, we observe $3 \times 3$ convolutions are generally more
compressible than $1 \times 1$ convolutions.
This is intuitive as $3 \times 3$ convolutions have more
parameters, thus leading to heavier redundancy.

\begin{table}[!t]
  \caption{
    Confusion matrices of the adjacent GroupConvs.
    Here, the neuron connectivity between ``Layer4-Bottleneck1-conv1'' and ``Layer4-Bottleneck1-conv2''
    of the ResNet-50-85\% model is demonstrated.
    Left: the learned neuron connectivity;
    Right: the neuron connectivity of the ShuffleNet \citep{zhang2018shufflenet}.
  }
  \begin{minipage}[!t]{0.48\textwidth}
  \newcommand{\CC}[1]{\cellcolor{gray!#1}}
  \renewcommand{\arraystretch}{1.15}
  \centering
  \setlength{\tabcolsep}{0.4mm}{
    \begin{tabular}{|l|c|c|c|c|c|c|c|c|}
      \hline
      & G1 & G2 & G3 & G4 & G5 & G6 & G7 & G8 \\\hline
      G1 & \CC{20} 6 & \CC{20} 6 & \CC{60} 10 & \CC{40} 8 & \CC{50} 9 & \CC{20} 6 & \CC{90} 13 & \CC{20} 6 \\\hline
      G2 & \CC{50} 9 & \CC{40} 8 & \CC{30} 7 & \CC{40} 9 & \CC{70} 11 & \CC{40} 8 & \CC{5} 4 & \CC{40} 8 \\\hline
      G3 & \CC{70} 11 & \CC{40} 8 & \CC{70} 11 & \CC{20} 6 & \CC{5} 4 & \CC{40} 8 & \CC{30} 7 & \CC{50} 9 \\\hline
      G4 & \CC{120} 16 & \CC{50} 9 & \CC{10} 5 & \CC{50} 9 & \CC{60} 10 & \CC{5} 4 & \CC{20} 6 & \CC{10} 5 \\\hline
      G5 & \CC{30} 7 & \CC{50} 9 & \CC{30} 7 & \CC{30} 7 & \CC{40} 8 & \CC{60} 10 & \CC{50} 9 & \CC{30} 7 \\\hline
      G6 & \CC{10} 5 & \CC{30} 7 & \CC{60} 10 & \CC{20} 6 & \CC{30} 7 & \CC{70} 11 & \CC{30} 7 & \CC{70} 11 \\\hline
      G7 & \CC{5}  4 & \CC{40} 8 & \CC{30} 7 & \CC{100} 14 & \CC{20} 6 & \CC{40} 8 & \CC{30} 7 & \CC{60} 10 \\\hline
      G8 & \CC{20} 6 & \CC{50} 9 & \CC{30} 7 & \CC{10} 5 & \CC{50} 9 & \CC{50} 9 & \CC{70} 11 & \CC{40} 8 \\\hline
    \end{tabular}}
  \end{minipage}
  \hfill
  \begin{minipage}[!t]{0.48\textwidth}
    \newcommand{\CC}[1]{\cellcolor{gray!#1}}
    \renewcommand{\arraystretch}{1.15}
    \centering
    \setlength{\tabcolsep}{0.4mm}{
      \begin{tabular}{|l|c|c|c|c|c|c|c|c|}
        \hline
        & G1 & G2 & G3 & G4 & G5 & G6 & G7 & G8 \\\hline
        G1 & \CC{40} 8 & \CC{40} 8 & \CC{40} 8 & \CC{40} 8 & \CC{40} 8 & \CC{40} 8 & \CC{40} 8 & \CC{40} 8 \\\hline
        G2 & \CC{40} 8 & \CC{40} 8 & \CC{40} 8 & \CC{40} 8 & \CC{40} 8 & \CC{40} 8 & \CC{40} 8 & \CC{40} 8 \\\hline
        G3 & \CC{40} 8 & \CC{40} 8 & \CC{40} 8 & \CC{40} 8 & \CC{40} 8 & \CC{40} 8 & \CC{40} 8 & \CC{40} 8 \\\hline
        G4 & \CC{40} 8 & \CC{40} 8 & \CC{40} 8 & \CC{40} 8 & \CC{40} 8 & \CC{40} 8 & \CC{40} 8 & \CC{40} 8 \\\hline
        G5 & \CC{40} 8 & \CC{40} 8 & \CC{40} 8 & \CC{40} 8 & \CC{40} 8 & \CC{40} 8 & \CC{40} 8 & \CC{40} 8 \\\hline
        G6 & \CC{40} 8 & \CC{40} 8 & \CC{40} 8 & \CC{40} 8 & \CC{40} 8 & \CC{40} 8 & \CC{40} 8 & \CC{40} 8 \\\hline
        G7 & \CC{40} 8 & \CC{40} 8 & \CC{40} 8 & \CC{40} 8 & \CC{40} 8 & \CC{40} 8 & \CC{40} 8 & \CC{40} 8 \\\hline
        G8 & \CC{40} 8 & \CC{40} 8 & \CC{40} 8 & \CC{40} 8 & \CC{40} 8 & \CC{40} 8 & \CC{40} 8 & \CC{40} 8 \\\hline
      \end{tabular}}
  \end{minipage}
  % }
  \label{appendix-tab:permutation}
\end{table}

Furthermore, we illustrate the learned neuron connectivity and compare with
the ShuffleNet~\citep{zhang2018shufflenet} counterpart.
Here, we consider the channel permutation between two group convolutions
(GroupConvs) and demonstrate the connectivity via the \textit{confusion matrix}.
Specifically, we assume the first GroupConv is of cardinality $G_1$ and the second
of $G_2$, then the confusion matrix $\mD$ is a $G_1 \times G_2$ matrix
with $D_{i,j}$ denoting the number of channels 
that come from the $i^{th}$ group of the first GroupConv and belong to
the $j^{th}$ group of the second.

\begin{figure*}[!t]
  \centering
  \begin{overpic}[width=0.9\textwidth]{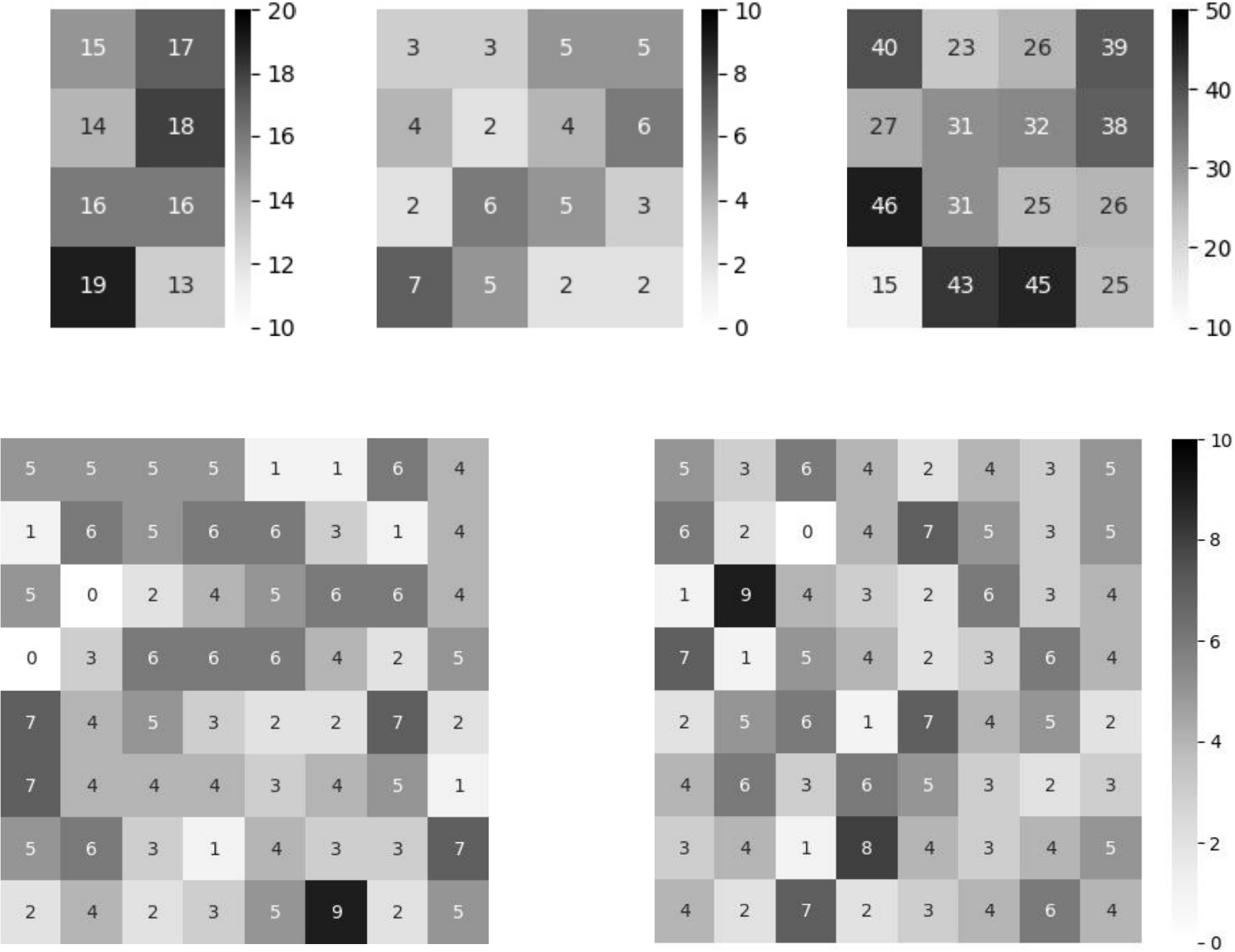}
    \put(0.0,47.0){(a) DenseNet-201-60\%}
    \put(-1.0,43.4){Block4-Layer24-conv1-2}
    \put(34.0,47.0){(b) ResNet-50-85\%}
    \put(29.0,43.4){Layer1-Bottleneck1-conv2-3}
    \put(72.0,47.0){(c) ResNet-101-80\%}
    \put(68.7,43.4){Layer4-Bottleneck2-conv2-3}
    \put(9.3,-2.5){(d) ResNet-50-85\%}
    \put(4.9,-6.1){Layer3-Bottleneck4-conv1-2}
    \put(64.3,-2.5){(e) ResNet-101-80\%}
    \put(59.9,-6.1){Layer3-Bottleneck1-conv1-2}
  \end{overpic}
  \vspace{20pt}
  \caption{More examples of the confusion matrices.}
  \vspace{-5pt}
  \label{fig:confusion}
\end{figure*}
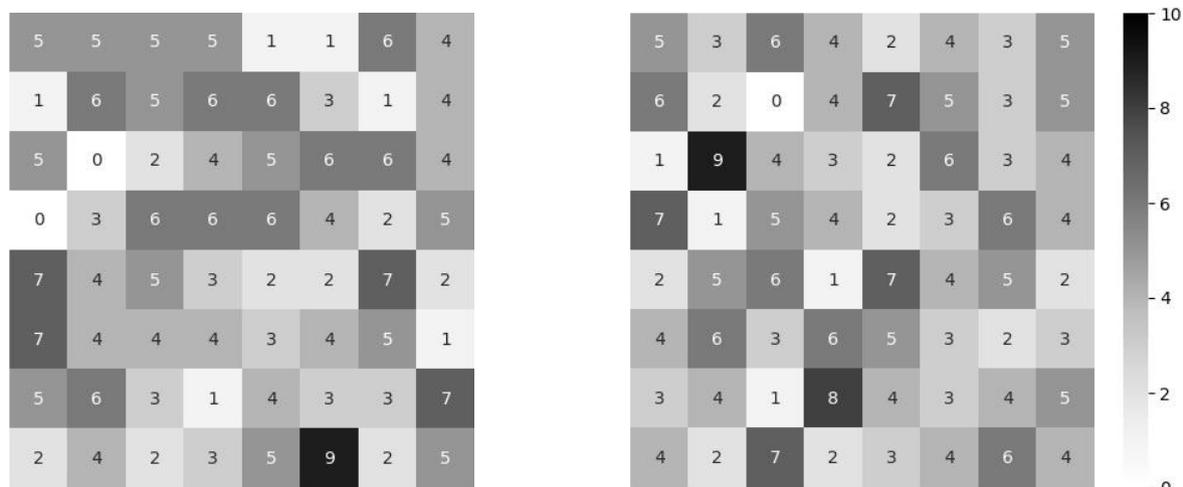

In \cref{appendix-tab:permutation}, we can see that the inter-group communication
is guaranteed as there are connections between every two groups.
Furthermore, the learnable channel shuffle scheme is more flexible.
The ShuffleNet~\citep{zhang2018shufflenet} scheme uniformly
partitions and distributes channels within each group,
while our approach allows small variations of the number of
connections for each group.
In this way, the network can itself control the information flow from each group
by customizing its neuron connectivity.
More examples can be found in \cref{fig:confusion}.
All models illustrated in this section are trained on the ImageNet dataset.

\end{document}